\documentclass[letterpaper]{article} 

\usepackage{aaai2027}  

\usepackage[hyphens]{url}  
\usepackage{graphicx} 
\usepackage{natbib}  
\usepackage{caption} 
\usepackage{multicol}
\usepackage{graphicx}
\usepackage{subcaption}
\usepackage{multirow}
\usepackage{array}
\usepackage{algorithm}
\usepackage{algorithmic}
\usepackage{diagbox}
\usepackage{booktabs} 
\usepackage{multirow}
\usepackage{enumitem}
\usepackage{amsmath,amsfonts,bm}
\usepackage{makecell}
\usepackage{amssymb}
\usepackage{tabularx}
\usepackage{graphicx}
\usepackage{caption}
\usepackage[table]{xcolor}
\newlength\savewidth
\newcommand{\shline}{\noalign{\global\savewidth\arrayrulewidth
  \global\arrayrulewidth 1pt}
  \hline
  \noalign{\global\arrayrulewidth\savewidth}}
\usepackage{cleveref}
\crefname{figure}{Figure}{Figure}
\crefname{table}{Table}{Table}
\crefname{section}{Section}{Section}
\crefname{appendix}{Appendix}{Appendix}
\usepackage{newfloat}
\usepackage{listings}
\DeclareCaptionStyle{ruled}{labelfont=normalfont,labelsep=colon,strut=off} 
\floatstyle{ruled}
\newfloat{listing}{tb}{lst}{}
\floatname{listing}{Listing}

\usepackage{booktabs}

\title{
Layer-Aware Position Embeddings for Visual Token Pruning \\
in Multimodal Large Language Models
}

\author{
    Yahong Wang\textsuperscript{\rm 1},
    Zhangkai Ni\textsuperscript{\rm 1}\corresponding,
    Juncheng Wu\textsuperscript{\rm 2},
    Yuyin Zhou\textsuperscript{\rm 2},
    Ying Wen\textsuperscript{\rm 3},
    Lianghua He\textsuperscript{\rm 1,4}\corresponding
}

\affiliations{
    \textsuperscript{\rm 1}Tongji University \quad
    \textsuperscript{\rm 2}University of California, Santa Cruz\\
    \textsuperscript{\rm 3}East China Normal University \quad
    \textsuperscript{\rm 4}Shanghai Eye Disease Prevention and Treatment Center \\
}

\begin{document}

\maketitle

\begin{abstract}
Multimodal large language models (MLLMs) incur substantial computational overhead due to the reliance on hundreds of visual tokens to represent images. 
While token pruning has emerged as a promising approach to reduce the inference cost of MLLMs, existing methods typically reassign position embeddings to the retained tokens using either sparse or continuous position embeddings, each introducing distinct limitations.
Sparse position embeddings tend to decrease the attention value allocated to visual tokens, thereby degrading the perception capability of MLLMs, 
whereas continuous position embeddings disrupt the original spatial correspondence of visual tokens, leading to weakened grounding capability.
To mitigate this issue, we perform layer-wise analysis of the language decoder and observe that intermediate layers play a critical role for maintaining the grounding capability of MLLMs under token pruning.
Based on this observation, we propose a layer-aware position embedding strategy, which switches to sparse position embeddings at grounding-sensitive layers while maintaining continuous position embeddings elsewhere. 
Extensive experiments across representative pruning methods and diverse benchmarks demonstrate that our approach improves the comprehensive multimodal performance of pruned MLLMs compared with standard sparse and continuous position embeddings.

\end{abstract}


\begin{links}
    \link{Code}{https://github.com/YahongWang1/LayerPos}
\end{links}

\begin{figure}[t]
\begin{center}
\centerline{\includegraphics[width=0.95\linewidth]{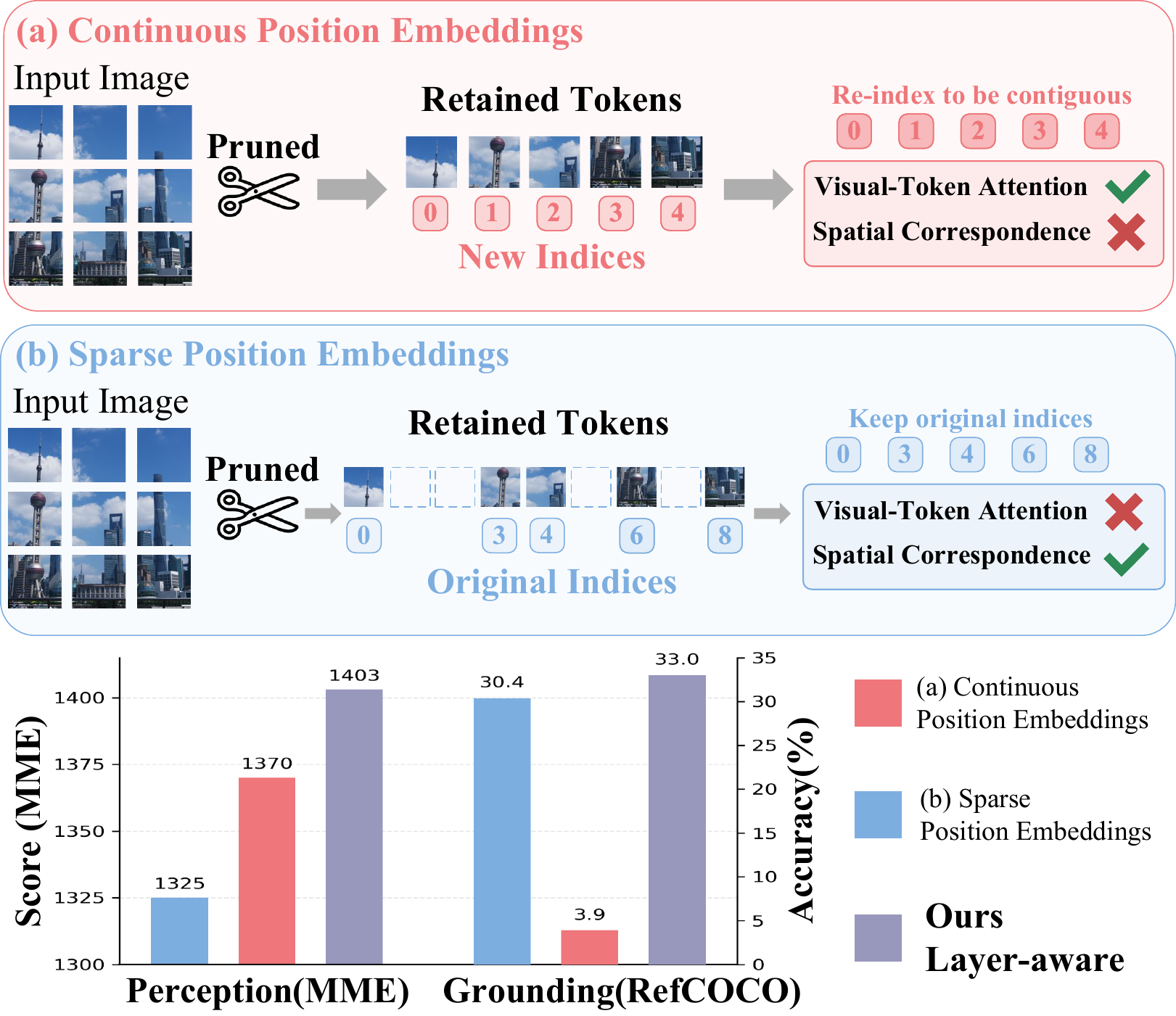}}
\caption{
\textbf{Position embedding trade-off after visual token pruning.}
(a) Continuous position embeddings re-index the position indices of retained visual tokens into a consecutive sequence, preserving MLLMs' attention to visual tokens but disrupting spatial correspondence, thereby degrading the grounding capability of pruned MLLMs.
(b) Sparse position embeddings preserve original positional indices and spatial correspondence, but large index gaps reduce MLLMs' attention to visual tokens and weaken the perception capability of pruned MLLMs.
Our method better balances perception and grounding performance.
}
\label{fig:teaser}
\end{center}
\end{figure}

\section{Introduction}

Multimodal large language models (MLLMs)~\citep{bai2025qwen2,li2024llava,wang2025internvl35,chen2024internvl} have achieved remarkable progress across a wide range of multimodal tasks, including general perception~\citep{fu2024mmecomprehensiveevaluationbenchmark,mmstar2024}, knowledge reasoning~\citep{lu2022learn,ai2d2016}, and visual grounding~\citep{refcoco,flickr30}. 
By integrating a visual encoder~\citep{radford2021learning,zhai2023sigmoid,chen2024internvl} with a Large Language Model (LLM)~\citep{vicuna2023,2023internlm,bai2023qwentechnicalreport}, these models enable general-purpose reasoning over multimodal inputs.
However, existing MLLMs usually convert images into extensive visual tokens, which dominate the input sequence length and significantly slow down the inference process. 
For example, LLaVA-1.5~\citep{liu2024improved} represents each image using 576 visual tokens, while InternVL3.5~\citep{wang2025internvl35} adopts a resolution-adaptive strategy that can produce several thousand visual tokens for high-resolution inputs.


Reducing the number of visual tokens is essential for efficient deployment of MLLMs, and existing training-free visual token pruning methods can be broadly categorized into two groups: 
(1) \textbf{pre-LLM pruning methods}, which removes redundant visual tokens before they are fed into the LLM~\citep{zhang2026beyond,alvar2025divprune,zooprune26cvpr,chen2026otprune}; 
(2) \textbf{intra-LLM pruning methods}, which prunes visual tokens within the layers of the LLM during inference~\citep{chen2024image,wen2025stop,wang2026entropyprune,zhang2024sparsevlm}.
These methods have demonstrated strong effectiveness in reducing inference cost while largely maintaining MLLM performance. 
However, they typically overlook an important factor in the pruning process: the design of position embeddings for the retained visual tokens. 
The top two panels of \cref{fig:teaser} illustrate the two position embedding strategies mainly adopted after token pruning.
\textbf{Sparse position embeddings} preserve the original position indices of retained visual tokens, resulting in large index gaps~\citep{chen2024image,wen2025stop}.
\textbf{Continuous position embeddings} re-index the position indices of retained tokens into a consecutive sequence, preserving positional continuity but distorting their original spatial correspondence~\citep{zhang2026beyond,alvar2025divprune,zooprune26cvpr}.
This distinction is critical because most MLLMs adopt rotary position embeddings (RoPE)~\citep{su2023rope}, where the attention between tokens is sensitive to their relative positional distances.
Sparse position embeddings introduce large gaps in the position indices, which reduces the attention allocated by MLLMs to visual tokens and thereby degrades their perception capability over visual content. 
Conversely, continuous position embeddings disrupt the original spatial relationships among visual tokens, substantially weakening the visual grounding capability of pruned MLLMs.
As shown in the bottom panel of \cref{fig:teaser}, sparse position embeddings lead to inferior performance on perception tasks, decreasing the MME score from 1370 to 1325~\citep{fu2024mmecomprehensiveevaluationbenchmark}. 
Conversely, continuous position embeddings cause a substantial degradation on grounding tasks, reducing the accuracy of RefCOCO\_testA from 30.4 to 3.9~\citep{refcoco}.
All results are obtained using LLaVA-1.5-7B with OTPrune~\citep{chen2026otprune}, where 77.8\% of visual tokens are pruned.

This trade-off raises a natural question: can we preserve positional continuity for perception while restoring spatial correspondence for grounding?
Inspired by~\citep{shi2026vision}, we hypothesize that intermediate decoder layers play a critical role in preserving the grounding capability of MLLMs.
To verify this hypothesis, we perform a layer-wise analysis under visual token pruning by selectively replacing continuous position embeddings with sparse position embeddings at different decoder layers.
Our analysis reveals that intermediate decoder layers are sensitive to sparse position embeddings, where applying sparse position embeddings significantly improves grounding performance.

Based on the above findings, we propose a layer-aware position embedding strategy. 
We identify a set of grounding-sensitive layers and switch their position embeddings to sparse position embeddings, while keeping continuous position embeddings in the remaining layers. 
This design preserves positional continuity for visual perception while restoring spatial correspondence at grounding-sensitive layers, thereby improving grounding capability without sacrificing perception performance.
Extensive experiments demonstrate that our layer-aware strategy consistently improves the performance of pruned MLLMs across a wide range of tasks, including visual perception, knowledge reasoning, relational understanding, and visual grounding, compared with both sparse and continuous position embeddings.
On LLaVA-1.5-7B pruned by CDPruner~\citep{zhang2026beyond} with 77.8\% visual tokens removed, our strategy preserves 98.2\% of the original model's performance, achieving 7.9\% and 13.3\% improvements over sparse and continuous position embeddings, respectively.
In addition, our strategy generalizes effectively to different MLLMs, including LLaVA-1.5-13B~\citep{liu2024improved}, a larger-scale MLLM, and InternVL3.5-8B~\citep{wang2025internvl35}, a cutting-edge open-source MLLM, highlighting its scalability and robustness. 
In summary, our contributions are as follows:
\begin{itemize}
\item We identify a critical yet overlooked trade-off in visual token pruning, showing that sparse and continuous position embeddings respectively lead to degraded perception and grounding capabilities.

\item Our empirical analysis reveals that intermediate decoder layers are critical for preserving grounding capability of MLLMs. Motivated by this finding, we propose a layer-aware position embedding strategy, which applies sparse and continuous position embeddings at different decoder layers in a structured manner.

\item We conduct extensive evaluations on diverse MLLM architectures and benchmarks, demonstrating consistent improvements over existing position embedding across a wide range of tasks.
\end{itemize}

\section{Related Work}
\label{sec:relatedWork}

\subsection{Visual Token Pruning in MLLMs}
To reduce the computational burden introduced by long visual sequences in MLLMs, visual token pruning has emerged as a promising acceleration strategy.
Among these approaches, training-free pruning methods have become one of the mainstream paradigms, owing to their plug-and-play nature and compatibility with off-the-shelf MLLMs.
Existing training-free methods can be broadly categorized into two paradigms according to where pruning is performed.
\textbf{Pre-LLM pruning methods} remove redundant visual tokens before they are fed into the language decoder~\citep{zhang2026beyond,zhang2026vispruner,alvar2025divprune,zooprune26cvpr,chen2026otprune}. 
These methods operate on the visual features produced by the vision encoder or the vision-language projector, selecting retained tokens based on [CLS]-token attention, token diversity, or task relevance.
However, a critical limitation of pre-LLM methods is their reliance on pre-LLM features, which may fail to leverage information within the LLM and thus lead to biased pruning decisions.
\textbf{Intra-LLM pruning methods} prune visual tokens within the LLM layers during inference~\citep{chen2024image,wen2025stop,wang2026entropyprune,zhang2024sparsevlm,wang_when_token}.
These methods dynamically remove less important visual tokens according to intermediate representations or attention-related signals, thereby leveraging information within the LLM to make more adaptive pruning decisions.
Although these paradigms strike a reasonable trade-off between inference speed and task accuracy, they typically assign position embeddings to the retained tokens in either a sparse or continuous manner, which respectively compromises the visual perception and spatial grounding capabilities of pruned MLLMs.

\subsection{Position Embeddings after Visual Token Pruning}
Existing MLLMs typically adopt rotary position embeddings (RoPE)~\citep{su2023rope} to encode positional information in both visual and textual tokens, owing to its strong extrapolation capability.
Prior studies have shown that directly applying continuous position embeddings to retained visual tokens can significantly degrade performance on visual grounding tasks~\citep{chien2025gap,sun2026ivcprune}. 
This is because re-indexing pruned tokens into a compact sequence disrupts their original spatial correspondence, weakening fine-grained region-level alignment.
Restore shows that sparse position embeddings can effectively recover grounding capability~\citep{cho2026restore}. 
However, this comes at the cost of reduced attention allocation from text tokens to visual tokens.
Despite these efforts, existing approaches still rely on a fixed position embedding strategy across all layers, failing to simultaneously balance perception and grounding performance. 
In contrast, we observe that different decoder layers exhibit distinct sensitivities to grounding tasks. 
Motivated by this observation, we propose a layer-aware position embedding strategy, enabling a better trade-off between visual perception and grounding capabilities in pruned MLLMs.

\section{Method}
\label{sec:method}

\subsection{Preliminaries}
\label{Preliminaries}
Existing Transformer-based architectures typically inject positional information into self-attention through Rotary Position Embedding (RoPE)~\cite{su2023rope}. 
Unlike additive position embeddings, RoPE encodes positional signals by applying a position-dependent rotation to query and key representations in a pairwise feature space.


Specifically, let $\mathbf{x}_m, \mathbf{x}_n \in \mathbb{R}^{d}$ denote the hidden states of tokens at positions $m$ and $n$ in a sequence $\mathbf{X}$, where $d$ is the dimension of hidden states.
For each attention head, the query and key vectors are projected into $\mathbb{R}^{d_h}$, where $d_h$ denotes the attention head dimension.
The position-aware query and key representations are formulated as:
\begin{equation}
\mathbf{q}_m = \mathcal{R}(m)\mathbf{W}_q \mathbf{x}_m, \quad 
\mathbf{k}_n = \mathcal{R}(n)\mathbf{W}_k \mathbf{x}_n,
\label{eq:rope_qk}
\end{equation}
where $\mathbf{W}_q, \mathbf{W}_k \in \mathbb{R}^{d_h \times d}$ are projection matrices, and $\mathcal{R}(m) \in \mathbb{R}^{d_h \times d_h}$ denotes a block-wise rotation operator parameterized by position $m$, defined as:
\begin{equation}
\begin{pmatrix}
\cos m\theta_1 & -\sin m\theta_1 & \cdots & 0 & 0 \\
\sin m\theta_1 & \cos m\theta_1 & \cdots & 0 & 0 \\
\vdots & \vdots & \ddots & \vdots  & \vdots \\
0 & 0 & \cdots & \cos m\theta_{d_h/2}  & -\sin m\theta_{d_h/2} \\
0 & 0 & \cdots & \sin m\theta_{d_h/2} & \cos m\theta_{d_h/2}
\end{pmatrix}
\label{eq:rope_matrix}
\end{equation}
where $\theta_i = 10000^{-2(i-1)/d_h}$ for $i \in \{1, 2, \ldots, d_h/2\}$.

Under this construction, the attention logit between token $m$ and $n$ can be written as:
\begin{equation}
z_{m,n} = \frac{(\mathbf{q}_m)^\top \mathbf{k}_n}{\sqrt{d_h}}
= \frac{(\mathbf{W}_q \mathbf{x}_m)^\top \mathcal{R}(n-m)\mathbf{W}_k \mathbf{x}_n}{\sqrt{d_h}}.
\label{eq:rope_attn}
\end{equation}
This formulation shows that the attention score is influenced by the relative distance between tokens $(m-n)$.
This property allows RoPE to implicitly encode relative positional relationships while preserving the norm of representations.

\subsection{Position Embedding Trade-off after Token Pruning}
\label{sec:position_embedding_tradeoff}

\begin{figure}[t]
    \centering
    \includegraphics[width=\linewidth]{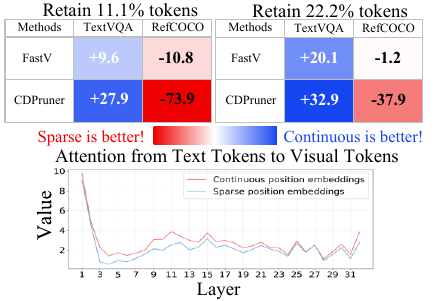}
    \caption{
    \textbf{Task-dependent effects of position embeddings after pruning.}
    \textbf{Top:} Performance gap between continuous and sparse position embeddings: continuous embeddings perform better on TextVQA, while sparse embeddings are more effective on RefCOCO.
    \textbf{Bottom:} Attention from text tokens to visual tokens on LLaVA-1.5-7B, pruned by CDPruner with 64 tokens retained.
    Continuous embeddings yield higher attention to visual tokens than sparse embeddings.
    }
    \label{fig:sparse_vs_con_vis_attn}
\end{figure}

After visual token pruning, the retained tokens can be assigned with different position embedding strategies.
A common choice is \textit{continuous position embeddings}, which re-indexes the position indices of retained tokens into a consecutive sequence.
Another choice is \textit{sparse position embeddings}, which preserves their original positional indices before pruning.
Although both strategies are simple, they induce different effects on pruned MLLMs.

To study this effect, we conduct experiments on LLaVA-1.5-7B with two representative pruning methods: FastV~\cite{chen2024image}, an intra-LLM pruning method, and CDPruner~\cite{zhang2026beyond}, a pre-LLM pruning method.
We evaluate two pruning ratios, retaining 64 visual tokens and 128 visual tokens, corresponding to 11.1\% and 22.2\% of the original 576 visual tokens, respectively.
For each setting, we compare sparse and continuous position embeddings on TextVQA~\cite{singh2019towards} and RefCOCO~\cite{refcoco}, which represent perception-oriented and grounding-oriented tasks, respectively.
As shown in \Cref{fig:sparse_vs_con_vis_attn}, each entry reports the performance gap between continuous and sparse position embeddings:
$\Delta = \mathrm{Acc}_{\mathrm{cont}} - \mathrm{Acc}_{\mathrm{sparse}}.$
Positive values indicate that continuous position embeddings perform better, while negative values indicate that sparse position embeddings perform better.

As shown in the top of \Cref{fig:sparse_vs_con_vis_attn}, the two position embedding strategies exhibit a clear task-dependent trade-off.
On RefCOCO, sparse position embeddings consistently outperform continuous position embeddings, indicating that preserving the original spatial indices of retained visual tokens is critical for grounding tasks.
In contrast, on TextVQA, continuous position embeddings consistently achieve better performance, suggesting that positional continuity is more beneficial for perception-oriented tasks.
Moreover, the advantage of continuous position embeddings is more pronounced for CDPruner than FastV.
This suggests that pre-LLM pruning methods are more sensitive to the position embedding strategy, since all decoder layers receive the pruned visual sequence from the beginning.
We also observe that the gap between continuous and sparse position embeddings on TextVQA becomes larger when fewer visual tokens are retained, showing that the position embedding issue is amplified under more aggressive pruning.

We further analyze this phenomenon through attention visualization.
Specifically, we sample 1,000 examples from TextVQA, feed them into LLaVA-1.5-7B pruned by CDPruner with 64 visual tokens retained, and compute the summed attention from all text tokens to the retained visual tokens at each language decoder layer.
As shown in the bottom of \Cref{fig:sparse_vs_con_vis_attn}, continuous position embeddings consistently yield higher attention values than sparse position embeddings across most layers.
With sparse position embeddings, the retained visual tokens preserve their original spatial indices, but large index gaps are introduced after pruning.
These gaps increase the relative positional distances between tokens and weaken the attention from text tokens to visual tokens.
In contrast, continuous position embeddings re-index both text tokens and retained visual tokens into a compact sequence, reducing positional gaps and increasing the model's attention to visual information.
As a result, pruned MLLMs can better perceive visual content, leading to improved performance on perception tasks.
However, this re-indexing also distorts the original spatial correspondence of visual tokens, leading to degraded grounding performance.

\begin{figure}
\centering
\includegraphics[width=0.48\textwidth]{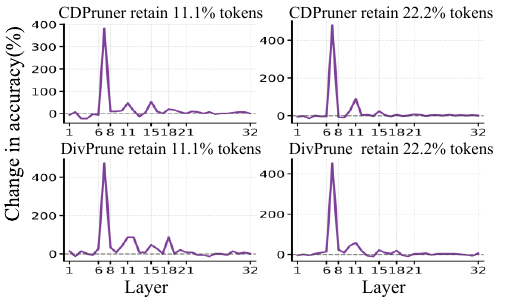}
\caption{
\textbf{Layer-wise grounding sensitivity to sparse position embeddings in pruned LLaVA-1.5-7B.}
Change in grounding accuracy when replacing continuous position embeddings with sparse position embeddings at each decoder layer.
A consistent layer-wise trend is observed across pruning methods and retention ratios, with grounding accuracy increasing sharply at intermediate layers and remaining near zero in deeper layers.
}
\label{fig:layer-wise}
\end{figure}

This observation can also be explained from the perspective of RoPE.
For RoPE-based MLLMs, the interaction between a query at position $m$ and a key at position $n$ depends on their relative position $m-n$.
Following RoPE~\cite{su2023rope}, the attention logit $z_{m,n}$ can be bounded by
\begin{equation}
\left(\max_i |h_{i+1}-h_i|\right)
\sum_{i=0}^{d_h/2-1}|S_{i+1}|,
\label{eq:rope_decay_bound}
\end{equation}
where 
$h_i = \mathbf{q}_{m,[2i:2i+1]}\mathbf{k}_{n,[2i:2i+1]}^{*}$ 
and $S_j=\sum_{i=0}^{j-1}e^{\mathrm{i}(m-n)\theta_i}$.
The normalized term $\frac{1}{d/2}\sum_{i=1}^{d/2}|S_i|$ decays as the relative distance $m-n$ increases~\cite{su2023rope}.
Therefore, sparse position embeddings, which introduce large positional gaps after pruning, tend to reduce token attention.
Continuous position embeddings alleviate this long-range decay by reducing positional gaps, but at the cost of distorting the original spatial correspondence among visual tokens.

These results reveal that neither sparse nor continuous position embeddings are universally optimal.
Inspired by prior findings that different layers in MLLMs may exhibit distinct functional roles~\cite{shi2026vision}, we hypothesize that intermediate decoder layers play a critical role in preserving the grounding capability of MLLMs.

\begin{figure*}[t]
\begin{center}
\centerline{\includegraphics[width=1.0\linewidth]{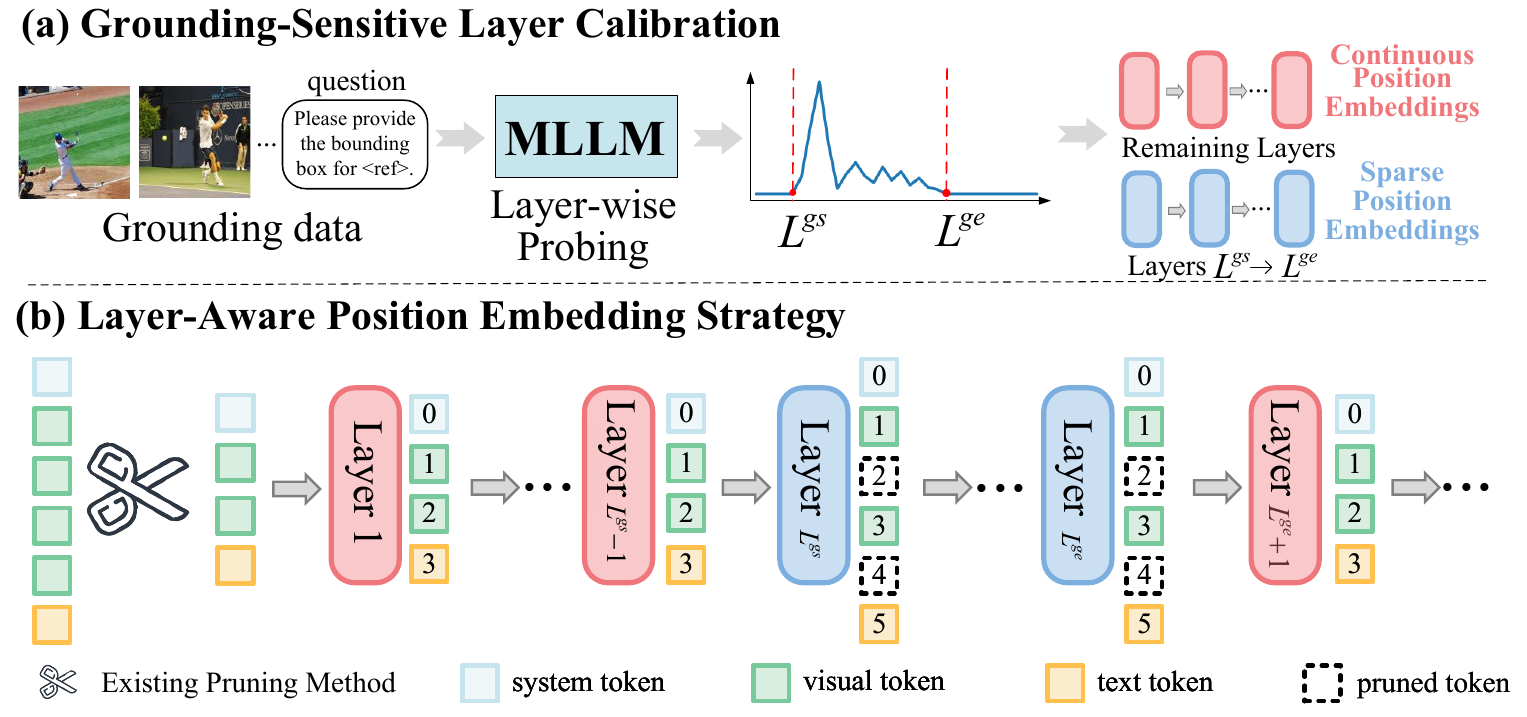}}
\caption{
\textbf{Overview of our method.}
(a) \textbf{Grounding-Sensitive Layer Calibration} locates the decoder-layer interval $[L^{gs}, L^{ge}]$ where grounding accuracy is most sensitive to sparse position embeddings.
(b) \textbf{Layer-Aware Position Embedding Strategy} applies sparse position embeddings within $[L^{gs}, L^{ge}]$ and continuous position embeddings in the remaining layers, achieving a better balance between perception and grounding capabilities after token pruning.
}
\label{fig:framework}
\end{center}
\end{figure*}

\subsection{Grounding Layer Calibration}

\label{sec:grounding_layer_calibration}

To validate whether intermediate decoder layers play a critical role in preserving the grounding capability of pruned MLLMs, we conduct a layer-wise probing analysis under visual token pruning, as illustrated in \Cref{fig:framework}(a).
Specifically, we conduct experiments on LLaVA-1.5-7B under different pruning methods and token retention ratios, using 1,000 samples from RefCOCO.
We use the setting where continuous position embeddings are applied to all decoder layers as the baseline.
Then, for each decoder layer, we selectively switch its position embedding strategy from continuous position embeddings to sparse position embeddings, while keeping the remaining layers unchanged.
We measure the grounding accuracy change relative to the baseline:
\begin{equation}
    \Delta_l =
    \frac{\mathrm{Acc}_{l}^{\mathrm{sparse}}-\mathrm{Acc}^{\mathrm{cont}}}
    {\mathrm{Acc}^{\mathrm{cont}}}
    \times 100\%.
\end{equation}
where $\mathrm{Acc}_{l}^{\mathrm{sparse}}$ denotes the grounding accuracy when only the $l$-th decoder layer uses sparse position embeddings, and $\mathrm{Acc}^{\mathrm{cont}}$ denotes the grounding accuracy when all decoder layers use continuous position embeddings.

As shown in \Cref{fig:layer-wise}, switching a single decoder layer to sparse position embeddings leads to a sharp increase in grounding accuracy starting from the 7th layer, where the relative improvement exceeds 400\% over the baseline.
The improvement then fluctuates across subsequent layers until around the 21st layer, beyond which the gain nearly vanishes.
This trend is consistent across different pruning methods and token retention ratios, suggesting that such phenomenon reflects an intrinsic property of MLLMs rather than an artifact of a specific pruning configuration.

We define $L^{gs}$ as the first layer where grounding accuracy improvement becomes significant, and $L^{ge}$ as the last layer before the improvement converges to near zero.
Decoder layers within this interval, \textit{i.e.}, $l \in [L^{gs}, L^{ge}]$, are regarded as grounding-sensitive layers, which play a critical role in maintaining the grounding capability of MLLMs under token pruning.

\subsection{Layer-Aware Position Embedding Strategy}
\label{sec:layer_aware_position_embedding}

Based on the grounding-sensitive layers, we propose a layer-aware position embedding strategy for pruned MLLMs.
Instead of applying a fixed position embedding strategy to all decoder layers, our method assigns different position embeddings according to the role of each layer.
Let $\mathcal{G}=\{l \mid L^{gs} \leq l \leq L^{ge}\}$ denote the set of grounding-sensitive layers, the position embeddings for the $l$-th decoder layer are assigned as:
\begin{equation}
    \mathbf{p}^{(l)} =
    \begin{cases}
        \mathbf{p}^{\mathrm{sparse}}, & l \in \mathcal{G}, \\
        \mathbf{p}^{\mathrm{cont}}, & l \notin \mathcal{G},
    \end{cases}
\end{equation}
where $\mathbf{p}^{\mathrm{sparse}}$ denotes sparse position embeddings and $\mathbf{p}^{\mathrm{cont}}$ denotes continuous position embeddings.

As shown in \Cref{fig:framework}(b), our strategy can be directly applied to both pre-LLM and intra-LLM pruning methods.
After pruning, only a subset of visual tokens is retained.
For layers before the grounding-sensitive layers, we use continuous position embeddings, where the position indices of system tokens, retained visual tokens and text tokens are re-indexed into a consecutive sequence.
This reduces positional gaps and helps the pruned MLLM maintain attention to visual information.
For grounding-sensitive layers $l \in \mathcal{G}$, we switch the position embeddings to sparse position embeddings, which preserve the original position indices of retained visual tokens before pruning.
This design restores the spatial correspondence of visual tokens and helps preserve the grounding capability of the pruned MLLM.
For layers after the grounding-sensitive interval, the position embeddings switch back to continuous position embeddings for the remaining layers.
Therefore, our method preserves positional continuity in most layers while restoring spatial correspondence at grounding-sensitive layers.
This design avoids the limitations of fixed sparse or fixed continuous position embeddings, enabling a better balance between perception and grounding capabilities.
Notably, the proposed strategy only changes the position embeddings used in different decoder layers and does not modify model parameters or the token pruning process, making it training-free and compatible with existing pruning methods.

\section{Experiment}
\label{sec:experiment}

\subsection{Experiment Setup}
\label{Exp_stt}

\renewcommand{\multirowsetup}{\centering}
\definecolor{mygray}{gray}{.92}
\begin{table*}[t]
    \centering
    \setlength{\tabcolsep}{4.5pt}
    \renewcommand{\arraystretch}{1.33}
    \footnotesize
	\centering
    \begin{tabular}{lc|ccc|cc|c|c|cc}
        \shline
        \multicolumn{2}{c|}{} &
        \multicolumn{3}{c|}{\textbf{Perception}} &
        \multicolumn{2}{c|}{\textbf{Knowledge}} &
        \multicolumn{1}{c|}{\textbf{Relation}} &
        \multicolumn{1}{c|}{\textbf{Grounding}} &
        \multicolumn{2}{c}{\textbf{Average}} \\
        \textbf{Method} &
        \textbf{\makecell{Position\\Embeddings}} &
        MME$^{P}$ &
        MMstar &
        MMVP &
        SQA$^{I}$ &
        AI2D &
        VSR &
        RefCOCO &
        \textbf{Acc.(\%)} &
        \textbf{Rel.(\%)} \\
        \shline
        LLaVA-1.5-7B & - & 1347 & 58.8 & 21.3 & 65.3 & 52.0 & 69.7 & 55.0 & 55.6 & 100.0 \\
        \hline
        \rowcolor{mygray} 
        \multicolumn{11}{c}{ \textit{Retain 128 Tokens} \ ($\downarrow 77.8\%$) } \\ 
        \hline 
        \multirow{3}{*}{\makecell{FastV\\\texttt{(ECCV24)}}}  
        & Continuous  & 1240 & 51.6 & 13.3 & 64.5 & 49.0 & 67.4 & 1.0 & 44.1 & 79.3 \\
        & Sparse & 1247 & \textbf{55.2} & \textbf{15.7} & 65.6 & 50.2 & 68.1 & 7.1 & 46.3 & 83.3 \\
        & Ours   & \textbf{1264} & \textbf{55.2} & 14.7 & \textbf{65.8} & \textbf{50.3} & \textbf{68.6} & \textbf{7.5} & \textbf{46.5} & \textbf{83.6} \\
       \hline 
        \multirow{3}{*}{\makecell{DivPrune\\\texttt{(CVPR25)}}}
       & Continuous & 1364 & 54.0 & 18.0 & 64.3 & 50.3 & 68.6 & 4.2 & 46.8 & 84.2 \\
        & Sparse & 1146 & 54.4 & 20.0 & 63.2 & 46.3 & 68.1 & 38.6 & 49.7 & 89.4 \\
        & Ours & \textbf{1414} & \textbf{56.8} & \textbf{22.0} & \textbf{65.4} & \textbf{52.2} & \textbf{69.5} & \textbf{40.1} & \textbf{53.8} & \textbf{96.8} \\
        \hline 
        \multirow{3}{*}{\makecell{CDPruner\\\texttt{(NIPS25)}}}
       & Continuous & 1340 & 56.0 & 18.7 & 64.2 & 51.4 & 67.9 & 5.4 & 47.2 & 84.9 \\
        & Sparse & 1151 & 56.0 & 17.3 & 62.4 & 46.4 & 67.7 & 44.4 & 50.2 & 90.3 \\
        & Ours & \textbf{1400} & \textbf{58.8} & \textbf{22.0} & \textbf{65.3} & \textbf{51.8} & \textbf{69.3} & \textbf{45.1} & \textbf{54.6} & \textbf{98.2} \\
       \hline 
       \multirow{3}{*}{\makecell{OTPrune\\\texttt{(CVPR26)}}} 
       & Continuous & 1370 & 56.8 & 19.3 & 65.3 & 50.8 & 68.2 & 3.1 & 47.4 & 85.3 \\
        & Sparse & 1325 & 56.8 & \textbf{22.7} & 64.6 & 48.9 & 68.6 & 21.3 & 49.9 & 89.7 \\
        & Ours & \textbf{1403} & \textbf{59.6} & 22.0 & \textbf{65.4} & \textbf{51.3} & \textbf{69.6} & \textbf{23.0} & \textbf{51.6} & \textbf{92.8} \\
       \hline 
       \multirow{3}{*}{\makecell{ZOOPrune\\\texttt{(CVPR26)}}}  
       & Continuous & 1368 & 54.4 & 17.3 & 64.4 & 50.9 & 68.4 & 4.5 & 46.9 & 84.4 \\
        & Sparse & 1146 & 51.6 & 17.3 & 63.9 & 46.6 & 68.2 & 36.1 & 48.7 & 87.6 \\
        & Ours & \textbf{1397} & \textbf{57.6} & \textbf{24.7} & \textbf{65.2} & \textbf{51.7} & \textbf{69.7} & \textbf{37.9} & \textbf{53.8} & \textbf{96.8} \\
       \shline 
       \end{tabular}
    \caption{\textbf{Performance of our layer-aware position embedding strategy on LLaVA-1.5-7B.} The vanilla number of vision tokens is 576. \textbf{Acc.} represents the average accuracy across all benchmarks. \textbf{Rel.} denotes the relative performance retained compared to the original model. The best performance is highlighted in \textbf{bold}.}
    \label{tab:llava15}
\end{table*}

\textbf{Base Models.}
We evaluate our method on three representative MLLMs with different characteristics: LLaVA-1.5-7B~\citep{liu2024improved}, a general-purpose MLLM; LLaVA-1.5-13B~\citep{liu2024improved}, a larger-scale MLLM; and InternVL3.5-8B~\citep{wang2025internvl35}, a cutting-edge open-source MLLM.

\noindent\textbf{Benchmarks.}
We conduct experiments on benchmarks covering four multimodal capabilities: 
\textbf{Visual Perception}, including MME$^{P}$~\citep{fu2024mmecomprehensiveevaluationbenchmark}, MMstar~\citep{mmstar2024}, and MMVP~\citep{tong2024mmvp}; 
\textbf{Knowledge Reasoning}, including SQA$^{I}$~\citep{lu2022learn} and AI2D~\citep{ai2d2016}; 
\textbf{Relation Understanding} , including VSR~\citep{vsr}; 
and \textbf{Visual Grounding}, including RefCOCO~\citep{refcoco}.
For MMstar, we use its subsets corresponding to coarse perception.
More details of the benchmarks are provided in Appendix~\ref{app:benchmark}.

\noindent\textbf{Pruning Methods.}
We evaluate our approach on several representative visual token pruning methods for MLLMs, including  FastV~\citep{chen2024image}, 
DivPrune~\citep{alvar2025divprune},
CDPruner~\citep{zhang2026beyond},
ZOOPrune~\citep{zooprune26cvpr},
OTPrune~\citep{chen2026otprune}.
More details are provided in Appendix~\ref{app:baseline}.
For each pruning method, we compare our layer-aware position embedding strategy with sparse and continuous position embeddings.

\subsection{Performance on LLaVA-1.5-7B}
\label{sec:llava15}
As shown in~\cref{tab:llava15}, our layer-aware position embedding strategy consistently improves the overall performance of pruned LLaVA-1.5-7B across diverse visual token pruning methods when retaining only 128 visual tokens.
Compared with continuous and sparse position embeddings, our method achieves the highest average accuracy and relative performance under all five pruning methods.
For example, when pruned with CDPruner at a retention of 128 tokens, our method achieves 54.6\% average accuracy and preserves 98.2\% of the original model performance, outperforming sparse position embeddings by 4.4\% and 7.9\%, respectively.
The improvements are observed across all evaluated multimodal capabilities.
For visual perception, our method substantially improves the performance of pruned LLaVA-1.5-7B on MME$^{P}$, MMstar, and MMVP under most pruning methods; for visual grounding, it consistently achieves the best performance on RefCOCO.
Moreover, the enhancements are not limited to perception and grounding capabilities.
Our method also boosts the knowledge reasoning and relation understanding capabilities of pruned LLaVA-1.5-7B, evidenced by its improved performance on SQA$^{I}$, AI2D, and VSR.
These results demonstrate that our layer-aware position embedding strategy effectively balances positional continuity and spatial correspondence, leading to comprehensive performance improvements for pruned LLaVA-1.5-7B rather than benefiting only a specific capability.

\begin{table}[t]
  \centering
  \footnotesize
  \setlength{\tabcolsep}{1.4pt}
  \renewcommand{\arraystretch}{1.33}
  \begin{tabular}{lc|
      >{\centering\arraybackslash}p{0.9cm}
      >{\centering\arraybackslash}p{0.7cm}
      >{\centering\arraybackslash}p{0.75cm}
      >{\centering\arraybackslash}p{0.65cm}
      >{\centering\arraybackslash}p{0.65cm}
      >{\centering\arraybackslash}p{0.65cm}
      }
      \shline
      \textbf{Method} &
      {\small\textbf{\makecell{Position\\Embeddings}}} &
      MME$^{P}$ &
      MMs. &
      MMV. &
      VSR &
      AI2D &
      \textbf{Acc.} \\
      \shline

      LLaVA-13B & - & 1406 & 61.2 & 32.7 & 73.1  & 58.7 & 59.2 \\

      \hline
      \rowcolor{mygray}
      \multicolumn{8}{c}{
          \textit{Retain 128 Tokens} \ ($\downarrow 77.8\%$)
      } \\
      \hline

      \multirow{3}{*}{\makecell{FastV\\\texttt{(ECCV24)}}}
      & Cont.   & 1277 & 57.2 & 23.3 & 70.9 & 55.2 & 54.1 \\
      & Sparse  & 1290 & 58.0 & 22.7 & 71.0 & 55.3 & 54.3 \\
      & Ours    & 1294 & 58.0 & 26.7 & 71.1 & 55.0 & \textbf{55.1} \\

      \hline

      \multirow{3}{*}{\makecell{DivPrune\\\texttt{(CVPR25)}}}
      & Cont.   & 1299 & 54.0 & 28.7 & 71.6 & 56.6 & 55.2 \\
      & Sparse  & 944  & 53.6 & 29.3 & 68.6 & 45.4 & 48.8 \\
      & Ours    & 1339 & 57.6 & 31.3 & 71.1 & 58.3 & \textbf{57.1} \\

      \hline

      \multirow{3}{*}{\makecell{CDPruner\\\texttt{(NIPS25)}}}
      & Cont.   & 1312 & 56.0 & 28.7 & 69.3 & 56.2 & 55.2 \\
      & Sparse  & 699  & 52.4 & 27.3 & 66.2 & 35.7 & 43.3 \\
      & Ours    & 1346 & 60.4 & 29.3 & 70.9 & 57.4 & \textbf{57.1} \\

      \hline

      \multirow{3}{*}{\makecell{OTPrune\\\texttt{(CVPR26)}}}
      & Cont.   & 1355 & 55.6 & 27.3 & 70.7 & 55.9 & 55.5 \\
      & Sparse  & 1337 & 60.0 & 30.7 & 71.1 & 55.5 & 56.8 \\
      & Ours    & 1370 & 59.2 & 32.0 & 71.1 & 56.7 & \textbf{57.5} \\

      \shline
  \end{tabular}
  \caption{
      \textbf{Performance of our layer-aware position embedding strategy
      on LLaVA-1.5-13B.} Cont. denotes continuous position embeddings. 
      MMs. denotes the MMstar benchmark, and MMV. denotes the MMVP benchmark.
  }
  \label{tab:llava1513b}
\end{table}

\subsection{Performance on LLaVA-1.5-13B}
\label{sec:llava1513b}
LLaVA-1.5-13B is a larger-scale MLLM with enhanced multimodal understanding capabilities.
To evaluate the scalability of our layer-aware position embedding strategy to larger MLLMs, we further conduct experiments on LLaVA-1.5-13B.
As shown in~\cref{tab:llava1513b}, our method consistently improves the performance of pruned LLaVA-1.5-13B across different visual token pruning methods when retaining only 128 visual tokens.
When applied to DivPrune, our method achieves 57.1\% average accuracy, outperforming continuous and sparse position embeddings by 1.9\% and 8.3\%, respectively.
These improvements are not limited to the overall performance but are consistently observed across individual multimodal benchmarks.
For instance, under CDPruner, our method achieves 60.4\% on the coarse-perception subset of MMstar, significantly exceeding continuous position embeddings~(56.0\%) and sparse position embeddings~(52.4\%).
These results demonstrate the scalability of our layer-aware position embedding strategy to larger-scale MLLM.

\begin{table}[t]
    \centering
    \setlength{\tabcolsep}{1.5pt}
    \renewcommand{\arraystretch}{1.33}
    \footnotesize
    \centering
    \begin{tabular}{lc|
        >{\centering\arraybackslash}p{0.8cm}
        >{\centering\arraybackslash}p{0.7cm}
        *{3}{>{\centering\arraybackslash}p{0.65cm}}
        >{\centering\arraybackslash}p{0.65cm}
        }
        \shline
        \textbf{Method} &
        {\small\textbf{\makecell{Position\\Embeddings}}} &
        MMV. &
        SQA$^{I}$ &
        VSR &
        Ref. &
        \textbf{Acc.} &
        \textbf{Rel.} \\
        \shline
        InternVL3.5 & - & 42.7 & 86.3 & 64.7 & 84.9 & 69.7 & 100.0 \\
        \hline
        \rowcolor{mygray} \multicolumn{8}{c}{ \textit{Retain 25\% Tokens} \ ($\downarrow 75.0\%$) } \\
        \hline
        \multirow{3}{*}{\makecell{FastV\\\texttt{(ECCV24)}}} 
        & Cont. & 23.3 & 83.9 & 67.5 & 29.2 & 51.0 & 73.2 \\
        & Sparse & 31.3 & 83.7 & 66.6 &  72.8 & 63.6 & 91.2 \\
        & Ours & 32.0 & 83.9 & 66.3 & 73.0 & \textbf{63.8} & \textbf{91.5} \\
        \hline
        \multirow{3}{*}{\makecell{DivPrune\\\texttt{(CVPR25)}}} 
        & Cont. & 30.0 & 80.7 & 64.5 & 18.0 & 48.3 & 69.3 \\
        & Sparse & 31.3 & 82.4 & 65.8 & 65.7 & 61.3 & 87.9 \\
        & Ours & 32.0 & 82.5 & 66.3 & 66.2 & \textbf{61.8} & \textbf{88.7} \\
        \hline 
        \multirow{3}{*}{\makecell{CDPruner\\\texttt{(NIPS25)}}}
        & Cont. & 27.3 & 80.3 & 62.9 & 20.2 & 47.7 & 68.4 \\
        & Sparse & 30.7  & 81.5  & 64.9 & 64.0 & 60.3 & 86.5 \\
        & Ours & 34.0 & 82.1 & 64.8 & 64.2 & \textbf{61.3} & \textbf{87.9} \\
        \shline
    \end{tabular}
    \caption{\textbf{Performance of our layer-aware position embedding strategy on InternVL3.5-8B.} MMV. and Ref. denote the MMVP and RefCOCO benchmarks, respectively. }
    \label{tab:internvl358b}
\end{table}

\subsection{Performance on InternVL3.5-8B}
\label{sec:internvl358b}
InternVL3.5 series adopts a more advanced MLLM architecture that supports Naive Dynamic Resolution to process images of arbitrary aspect ratios.
To verify the architectural robustness of our method, we further conduct experiments on InternVL3.5-8B.
As shown in~\cref{tab:internvl358b}, our layer-aware position embedding strategy consistently achieves the highest average accuracy and relative performance across FastV, DivPrune, and CDPruner when retaining only 25\% of visual tokens.
Specifically, when combined with DivPrune, our method achieves 61.8\% average accuracy and retains 88.7\% of the original model performance, improving over continuous position embeddings by 12.5\% and 19.4\%, respectively.
The enhancements in multimodal capabilities span visual perception, knowledge reasoning, relation understanding, and visual grounding, as reflected by improved performance on MMVP, SQA$^{I}$, VSR, and RefCOCO, respectively.
Notably, our method even surpasses the base model on VSR by 1.6\%, when combined with DivPrune and retaining only 25\% of visual tokens. 
These results demonstrate the effectiveness and generalizability of our layer-aware position embedding strategy across diverse model architectures.

\section{Conclusion}
\label{sec:conclusion}

In this paper, we investigate the overlooked impact of position embeddings on visual token pruning in MLLMs. 
We reveal that sparse and continuous position embeddings introduce a fundamental trade-off between visual perception and grounding capabilities, making neither global optimal after token pruning. 
Through layer-wise analysis, we identify that intermediate decoder layers play a critical role in preserving visual grounding ability of pruned MLLMs. 
Based on this observation, we propose a layer-aware position embedding strategy that dynamically assigns sparse and continuous position embeddings according to the functional roles of different decoder layers. 
Extensive experiments across various pruning methods, MLLM architectures, and multimodal benchmarks demonstrate that our approach consistently improves the overall performance of pruned MLLMs while maintaining inference efficiency. 
We hope this work provides a new perspective on position embedding design for efficient multimodal large language models.

\section{Limitations}
\label{sec:limitation}
Although our layer-aware position embedding strategy consistently improves the performance of pruned MLLMs, several limitations remain. 
First, our method relies on a layer-wise sensitivity analysis to identify grounding-sensitive layers, which introduces additional evaluation costs before deployment. 
Although this calibration is performed only once for each MLLM and remains training-free, developing a more efficient layer selection mechanism is an interesting direction for future work.
Second, our experiments mainly focus on image-based MLLMs, while extending the proposed strategy to video MLLMs with longer temporal sequences and complex spatio-temporal token interactions requires further investigation.
Finally, our approach is designed for RoPE-based MLLMs, and extending position embedding strategies after visual token pruning to MLLMs with other positional encoding mechanisms remains an open research direction.


\bibliography{aaai2027}

@article{vsr,
    title = "Visual Spatial Reasoning",
    author = "Liu, Fangyu  and
      Emerson, Guy  and
      Collier, Nigel",
    journal = "Transactions of the Association for Computational Linguistics",
    volume = "11",
    year = "2023",
    address = "Cambridge, MA",
    publisher = "MIT Press",
    url = "https://aclanthology.org/2023.tacl-1.37/",
    doi = "10.1162/tacl_a_00566",
    pages = "635--651"
}

@InProceedings{tong2024mmvp,
    author    = {Tong, Shengbang and Liu, Zhuang and Zhai, Yuexiang and Ma, Yi and LeCun, Yann and Xie, Saining},
    title     = {Eyes Wide Shut? Exploring the Visual Shortcomings of Multimodal LLMs},
    booktitle = {Proceedings of the IEEE/CVF Conference on Computer Vision and Pattern Recognition (CVPR)},
    month     = {June},
    year      = {2024},
    pages     = {9568-9578}
}

@misc{chien2025gap,
      title={Grounding-Aware Token Pruning: Recovering from Drastic Performance Drops in Visual Grounding Caused by Pruning}, 
      author={Tzu-Chun Chien and Chieh-Kai Lin and Shiang-Feng Tsai and Ruei-Chi Lai and Hung-Jen Chen and Min Sun},
      year={2025},
      eprint={2506.21873},
      archivePrefix={arXiv},
      primaryClass={cs.CV},
      url={https://arxiv.org/abs/2506.21873}, 
}

@inproceedings{
shi2026vision,
title={Vision Function Layer in Multimodal {LLM}s},
author={Cheng Shi and Yizhou Yu and Sibei Yang},
booktitle={The Thirty-ninth Annual Conference on Neural Information Processing Systems},
year={2026},
url={https://openreview.net/forum?id=nTc0LSqtqE}
}

@inproceedings{
sun2026ivcprune,
title={{IVC}-Prune: Revealing the Implicit Visual Coordinates in {LVLM}s for Vision Token Pruning},
author={Zhichao Sun and Yidong Ma and Gang Liu and Nemo Chen and Xu Tang and Yao Hu and Yongchao Xu},
booktitle={The Fourteenth International Conference on Learning Representations},
year={2026},
url={https://openreview.net/forum?id=46LbXtFgBm}
}

@inproceedings{
cho2026restore,
title={Improving Visual Token Reduction via Rectifying Distortions for Efficient Multimodal {LLM} Inference},
author={Hyeonwoo Cho and Donghyeon Baek and Yewon Kim and Bumsub Ham},
booktitle={Forty-third International Conference on Machine Learning},
year={2026},
url={https://openreview.net/forum?id=QxU5sv16tg}
}

@misc{chen2026otprune,
      title={OTPrune: Distribution-Aligned Visual Token Pruning via Optimal Transport}, 
      author={Xiwen Chen and Wenhui Zhu and Gen Li and Xuanzhao Dong and Yujian Xiong and Hao Wang and Peijie Qiu and Qingquan Song and Zhipeng Wang and Shao Tang and Yalin Wang and Abolfazl Razi},
      year={2026},
      eprint={2602.20205},
      archivePrefix={arXiv},
      primaryClass={cs.CV},
      url={https://arxiv.org/abs/2602.20205}, 
}

@InProceedings{wang_when_token,
    author    = {Wang, Yahong and Wu, Juncheng and Ni, Zhangkai and Yang, Longzhen and Liu, Yihang and Yang, Chengmei and Wen, Ying and He, Lianghua and Tang, Xianfeng and Liu, Hui and Zhou, Yuyin},
    title     = {When Token Pruning is Worse than Random: Understanding Visual Token Information in VLLMs},
    booktitle = {Proceedings of the IEEE/CVF Conference on Computer Vision and Pattern Recognition (CVPR)},
    month     = {June},
    year      = {2026},
    pages     = {31910-31919}
}

@misc{su2023rope,
      title={RoFormer: Enhanced Transformer with Rotary Position Embedding}, 
      author={Jianlin Su and Yu Lu and Shengfeng Pan and Ahmed Murtadha and Bo Wen and Yunfeng Liu},
      year={2023},
      eprint={2104.09864},
      archivePrefix={arXiv},
      primaryClass={cs.CL},
      url={https://arxiv.org/abs/2104.09864}, 
}

@InProceedings{flickr30,
author = {Plummer, Bryan A. and Wang, Liwei and Cervantes, Chris M. and Caicedo, Juan C. and Hockenmaier, Julia and Lazebnik, Svetlana},
title = {Flickr30k Entities: Collecting Region-to-Phrase Correspondences for Richer Image-to-Sentence Models},
booktitle = {Proceedings of the IEEE International Conference on Computer Vision (ICCV)},
month = {December},
year = {2015}
}

@misc{wang2026entropyprune,
      title={EntropyPrune: Matrix Entropy Guided Visual Token Pruning for Multimodal Large Language Models}, 
      author={Yahong Wang and Juncheng Wu and Zhangkai Ni and Chengmei Yang and Yihang Liu and Longzhen Yang and Yuyin Zhou and Ying Wen and Lianghua He},
      year={2026},
      eprint={2602.17196},
      archivePrefix={arXiv},
      primaryClass={cs.CV},
      url={https://arxiv.org/abs/2602.17196}, 
}

@InProceedings{zooprune26cvpr,
    author    = {Kim, Youngeun and Zhang, Youjia and Liu, Huiling and Jung, Aecheon and Lee, Sunwoo and Hong, Sungeun},
    title     = {ZOO-Prune: Training-Free Token Pruning via Zeroth-Order Gradient Estimation in Vision-Language Models},
    booktitle = {Proceedings of the IEEE/CVF Conference on Computer Vision and Pattern Recognition (CVPR)},
    month     = {June},
    year      = {2026},
    pages     = {39572-39582}
}

@misc{wang2025internvl35,
      title={InternVL3.5: Advancing Open-Source Multimodal Models in Versatility, Reasoning, and Efficiency}, 
      author={Weiyun Wang and Zhangwei Gao and Lixin Gu and Hengjun Pu and others},
      year={2025},
      eprint={2508.18265},
      archivePrefix={arXiv},
      primaryClass={cs.CV},
      url={https://arxiv.org/abs/2508.18265}, 
}

@misc{bai2025qwen2,
      title={Qwen2.5-VL Technical Report}, 
      author={Shuai Bai and Keqin Chen and Xuejing Liu and Jialin Wang and Wenbin Ge and Sibo Song and Kai Dang and Peng Wang and Shijie Wang and Jun Tang and Humen Zhong and Yuanzhi Zhu and Mingkun Yang and Zhaohai Li and Jianqiang Wan and Pengfei Wang and Wei Ding and Zheren Fu and Yiheng Xu and Jiabo Ye and Xi Zhang and Tianbao Xie and Zesen Cheng and Hang Zhang and Zhibo Yang and Haiyang Xu and Junyang Lin},
      year={2025},
      eprint={2502.13923},
      archivePrefix={arXiv},
      primaryClass={cs.CV},
      url={https://arxiv.org/abs/2502.13923}, 
}

@misc{li2024llava,
      title={LLaVA-OneVision: Easy Visual Task Transfer}, 
      author={Bo Li and Yuanhan Zhang and Dong Guo and Renrui Zhang and Feng Li and Hao Zhang and Kaichen Zhang and Peiyuan Zhang and Yanwei Li and Ziwei Liu and Chunyuan Li},
      year={2024},
      eprint={2408.03326},
      archivePrefix={arXiv},
      primaryClass={cs.CV},
      url={https://arxiv.org/abs/2408.03326}, 
}

@article{lu2022learn,
  title={Learn to explain: Multimodal reasoning via thought chains for science question answering},
  author={Lu, Pan and Mishra, Swaroop and Xia, Tanglin and Qiu, Liang and Chang, Kai-Wei and Zhu, Song-Chun and Tafjord, Oyvind and Clark, Peter and Kalyan, Ashwin},
  journal={Advances in Neural Information Processing Systems},
  volume={35},
  pages={2507--2521},
  year={2022}
}

@misc{fu2024mmecomprehensiveevaluationbenchmark,
      title={MME: A Comprehensive Evaluation Benchmark for Multimodal Large Language Models}, 
      author={Chaoyou Fu and Peixian Chen and Yunhang Shen and Yulei Qin and Mengdan Zhang and Xu Lin and Jinrui Yang and Xiawu Zheng and Ke Li and Xing Sun and Yunsheng Wu and Rongrong Ji},
      year={2024},
      eprint={2306.13394},
      archivePrefix={arXiv},
      primaryClass={cs.CV},
      url={https://arxiv.org/abs/2306.13394}, 
}

@inproceedings{singh2019towards,
  title={Towards vqa models that can read},
  author={Singh, Amanpreet and Natarajan, Vivek and Shah, Meet and Jiang, Yu and Chen, Xinlei and Batra, Dhruv and Parikh, Devi and Rohrbach, Marcus},
  booktitle={Proceedings of the IEEE/CVF conference on computer vision and pattern recognition},
  pages={8317--8326},
  year={2019}
}

@inproceedings{chen2024image,
  title={An image is worth 1/2 tokens after layer 2: Plug-and-play inference acceleration for large vision-language models},
  author={Chen, Liang and Zhao, Haozhe and Liu, Tianyu and Bai, Shuai and Lin, Junyang and Zhou, Chang and Chang, Baobao},
  booktitle={European Conference on Computer Vision},
  pages={19--35},
  year={2024},
  organization={Springer}
}

@inproceedings{zhang2024sparsevlm,
  title={SparseVLM: Visual Token Sparsification for Efficient Vision-Language Model Inference},
  author={Zhang, Yuan and Fan, Chun-Kai and Ma, Junpeng and Zheng, Wenzhao and Huang, Tao and Cheng, Kuan and Gudovskiy, Denis and Okuno, Tomoyuki and Nakata, Yohei and Keutzer, Kurt and others},
  booktitle={International Conference on Machine Learning},
  year={2025}
}

@inproceedings{alvar2025divprune,
  title={Divprune: Diversity-based visual token pruning for large multimodal models},
  author={Alvar, Saeed Ranjbar and Singh, Gursimran and Akbari, Mohammad and Zhang, Yong},
  booktitle={Proceedings of the Computer Vision and Pattern Recognition Conference},
  pages={9392--9401},
  year={2025}
}

@inproceedings{wen2025stop,
    title = "Stop Looking for ``Important Tokens'' in Multimodal Language Models: Duplication Matters More",
    author = "Wen, Zichen  and
      Gao, Yifeng  and
      Wang, Shaobo  and
      Zhang, Junyuan  and
      Zhang, Qintong  and
      Li, Weijia  and
      He, Conghui  and
      Zhang, Linfeng",
    editor = "Christodoulopoulos, Christos  and
      Chakraborty, Tanmoy  and
      Rose, Carolyn  and
      Peng, Violet",
    booktitle = "Proceedings of the 2025 Conference on Empirical Methods in Natural Language Processing",
    month = nov,
    year = "2025",
    address = "Suzhou, China",
    publisher = "Association for Computational Linguistics",
    url = "https://aclanthology.org/2025.emnlp-main.505/",
    doi = "10.18653/v1/2025.emnlp-main.505",
    pages = "9961--9980",
    ISBN = "979-8-89176-332-6"
}

@inproceedings{liu2024improved,
  title={Improved baselines with visual instruction tuning},
  author={Liu, Haotian and Li, Chunyuan and Li, Yuheng and Lee, Yong Jae},
  booktitle={Proceedings of the IEEE/CVF conference on computer vision and pattern recognition},
  pages={26296--26306},
  year={2024}
}

@inproceedings{zhai2023sigmoid,
  title={Sigmoid loss for language image pre-training},
  author={Zhai, Xiaohua and Mustafa, Basil and Kolesnikov, Alexander and Beyer, Lucas},
  booktitle={Proceedings of the IEEE/CVF international conference on computer vision},
  pages={11975--11986},
  year={2023}
}

@inproceedings{radford2021learning,
  title={Learning transferable visual models from natural language supervision},
  author={Radford, Alec and Kim, Jong Wook and Hallacy, Chris and Ramesh, Aditya and Goh, Gabriel and Agarwal, Sandhini and Sastry, Girish and Askell, Amanda and Mishkin, Pamela and Clark, Jack and others},
  booktitle={International conference on machine learning},
  pages={8748--8763},
  year={2021},
  organization={PmLR}
}

@article{zhang2026beyond,
  title={Beyond attention or similarity: Maximizing conditional diversity for token pruning in mllms},
  author={Zhang, Qizhe and Liu, Mengzhen and Li, Lichen and Lu, Ming and Zhang, Yuan and Pan, Junwen and She, Qi and Zhang, Shanghang},
  journal={Advances in Neural Information Processing Systems},
  volume={38},
  pages={25438--25468},
  year={2026}
}

@InProceedings{zhang2026vispruner,
    author    = {Zhang, Qizhe and Cheng, Aosong and Lu, Ming and Zhang, Renrui and Zhuo, Zhiyong and Cao, Jiajun and Guo, Shaobo and She, Qi and Zhang, Shanghang},
    title     = {Beyond Text-Visual Attention: Exploiting Visual Cues for Effective Token Pruning in VLMs},
    booktitle = {Proceedings of the IEEE/CVF International Conference on Computer Vision (ICCV)},
    month     = {October},
    year      = {2025},
    pages     = {20857-20867}
}

@inproceedings{chen2024internvl,
  title={Internvl: Scaling up vision foundation models and aligning for generic visual-linguistic tasks},
  author={Chen, Zhe and Wu, Jiannan and Wang, Wenhai and Su, Weijie and Chen, Guo and Xing, Sen and Zhong, Muyan and Zhang, Qinglong and Zhu, Xizhou and Lu, Lewei and others},
  booktitle={Proceedings of the IEEE/CVF Conference on Computer Vision and Pattern Recognition},
  pages={24185--24198},
  year={2024}
}

@misc{vicuna2023,
    title = {Vicuna: An Open-Source Chatbot Impressing GPT-4 with 90\%* ChatGPT Quality},
    url = {https://lmsys.org/blog/2023-03-30-vicuna/},
    author = {Chiang, Wei-Lin and Li, Zhuohan and Lin, Zi and Sheng, Ying and Wu, Zhanghao and Zhang, Hao and Zheng, Lianmin and Zhuang, Siyuan and Zhuang, Yonghao and Gonzalez, Joseph E. and Stoica, Ion and Xing, Eric P.},
    month = {March},
    year = {2023}
}

@misc{bai2023qwentechnicalreport,
      title={Qwen Technical Report}, 
      author={Jinze Bai and Shuai Bai and Yunfei Chu and Zeyu Cui and Kai Dang and Xiaodong Deng and Yang Fan and Wenbin Ge and Yu Han and Fei Huang and Binyuan Hui and others},
      year={2023},
      eprint={2309.16609},
      archivePrefix={arXiv},
      primaryClass={cs.CL},
      url={https://arxiv.org/abs/2309.16609}, 
}

@misc{2023internlm,
    title={InternLM: A Multilingual Language Model with Progressively Enhanced Capabilities},
    author={InternLM Team},
    howpublished = {\url{https://github.com/InternLM/InternLM-techreport}},
    year={2023}
}

@misc{mmstar2024,
      title={Are We on the Right Way for Evaluating Large Vision-Language Models?}, 
      author={Lin Chen and Jinsong Li and Xiaoyi Dong and Pan Zhang and Yuhang Zang and Zehui Chen and Haodong Duan and Jiaqi Wang and Yu Qiao and Dahua Lin and Feng Zhao},
      year={2024},
      eprint={2403.20330},
      archivePrefix={arXiv},
      primaryClass={cs.CV},
      url={https://arxiv.org/abs/2403.20330}, 
}

@inproceedings{ai2d2016,
author="Kembhavi, Aniruddha
and Salvato, Mike
and Kolve, Eric
and Seo, Minjoon
and Hajishirzi, Hannaneh
and Farhadi, Ali",
editor="Leibe, Bastian
and Matas, Jiri
and Sebe, Nicu
and Welling, Max",
title="A Diagram is Worth a Dozen Images",
booktitle="Computer Vision -- ECCV 2016",
year="2016",
publisher="Springer International Publishing",
address="Cham",
pages="235--251",
isbn="978-3-319-46493-0"
}

@InProceedings{refcoco,
author="Yu, Licheng
and Poirson, Patrick
and Yang, Shan
and Berg, Alexander C.
and Berg, Tamara L.",
editor="Leibe, Bastian
and Matas, Jiri
and Sebe, Nicu
and Welling, Max",
title="Modeling Context in Referring Expressions",
booktitle="Computer Vision -- ECCV 2016",
year="2016",
publisher="Springer International Publishing",
address="Cham",
pages="69--85",
isbn="978-3-319-46475-6"
}


\clearpage
\twocolumn[%
\vbox{%
  \hsize\textwidth%
  \linewidth\hsize%
  \vskip 0.625in minus 0.125in%
  \centering%
  {\LARGE\bf Supplementary Material \par}%
  \vskip 1em plus 2fil%
}%
]

\appendix

\section{Detailed Experiment Settings}
\label{app:detail_exp}

\subsection{Benchmarks}
\label{app:benchmark}

\noindent\textbf{Visual Perception.} 
\\
\textbf{MME$^{P}$}~\citep{fu2024mmecomprehensiveevaluationbenchmark}. MME is a comprehensive evaluation benchmark consisting of 14 subtasks divided into perception and cognition categories. We adopt its perception subset, denoted as MME$^{P}$, which evaluates fundamental visual perception capabilities such as object existence, counting, and scene recognition.
\\ 
\textbf{MMstar}~\citep{mmstar2024}. MMstar is a vision-indispensable benchmark constructed by filtering out samples that can be answered without relying on visual inputs. It covers six capability dimensions and emphasizes challenging samples that require genuine visual understanding. We use its coarse-perception subsets to evaluate the coarse-perception capability of MLLMs. 
\\ 
\textbf{MMVP}~\citep{tong2024mmvp}. MMVP is designed to expose the visual shortcomings of MLLMs through carefully constructed image pairs containing subtle visual differences. Each pair is associated with questions that require models to distinguish fine-grained visual patterns that are often overlooked by conventional vision encoders. 
\\ 
\noindent\textbf{Knowledge Reasoning.}
\\
\textbf{ScienceQA} (\textbf{SQA$^{I}$})~\citep{lu2022learn}. SQA$^{I}$ is a dataset containing multimodal science questions annotated with detailed explanations. It evaluates the model's understanding of scientific concepts from different fields.
\\ 
\textbf{AI2D}~\citep{ai2d2016}. AI2D consists of diagrams and corresponding questions, challenging models to parse and reason about diagrammatic structures, arrows, and labels to solve geometry and science problems. 
\\ 
\noindent\textbf{Relation Understanding.} 
\\
\textbf{VSR}~\citep{vsr}. VSR is a visual spatial reasoning benchmark containing natural text-image pairs that cover diverse spatial relations, such as \textit{under}, \textit{right of}, and \textit{facing}. It requires models to determine whether a textual statement correctly describes the spatial relationship between objects in an image. We use its zero-shot split for evaluation.
\\ 
\noindent\textbf{Visual Grounding.} 
\\
\textbf{RefCOCO}~\citep{refcoco}. RefCOCO is a referring expression comprehension benchmark built upon images from MSCOCO. Given an image and a natural-language expression referring to a specific object, a model is required to localize the target object using its bounding box, thereby evaluating fine-grained cross-modal alignment and spatial grounding capability.

\subsection{Baselines}
\label{app:baseline}
\textbf{FastV}~\citep{chen2024image} leverages the attention value of the last text token to rank the information of visual tokens after the third layer.
\\
\textbf{DivPrune}~\citep{alvar2025divprune} utilizes a diversity-based pruning metric. Instead of relying on attention scores, it selects a subset of tokens that maximize the diversity of information, ensuring that the retained tokens cover the most distinct visual features.
\\
\textbf{CDPruner}~\citep{zhang2026beyond} reformulates token pruning via determinantal point process (DPP) to maximize the conditional diversity of retained tokens based on instruction relevance.
\\
\textbf{ZOOPrune}~\citep{zooprune26cvpr} estimates the importance of visual tokens through zeroth-order perturbations at the lightweight projection layer. It retains tokens whose perturbations exert a greater influence on the projected visual features, thereby preserving informative and complementary visual cues without requiring backpropagation.
\\
\textbf{OTPrune}~\citep{chen2026otprune} formulates visual token pruning as a distribution alignment problem via optimal transport. It selects a subset of tokens by minimizing the 2-Wasserstein distance between the original and pruned token distributions, preserving both local diversity and global representativeness.
\\

\subsection{Implementation Details.}
All experiments are conducted on Nvidia
A6000 GPU. The implementation is carried
out in Python 3.10, utilizing PyTorch 2.1.2, and
CUDA 11.8. All baseline settings follow the
original paper.

\section{Grounding-Sensitive Layers of MLLMs}
\label{app:grounding_layers}

We calibrate the grounding-sensitive layers for each MLLM following the layer-wise analysis described in Section~\ref{sec:grounding_layer_calibration}.
Specifically, we identify the range of decoder layers in which switching from continuous to sparse position embeddings consistently improves the grounding capability of pruned MLLMs. The calibrated grounding-sensitive layers for different MLLMs are summarized in Table~\ref{tab:grounding_layers}.

\begin{table}[t]
    \centering
    \small
    \setlength{\tabcolsep}{6pt}
    \renewcommand{\arraystretch}{1.15}
    \begin{tabular}{lc}
        \hline
        \textbf{MLLM (Total Layers)} & \textbf{Grounding-Sensitive Layers} \\
        \hline
        LLaVA-1.5-7B (32)  & 7--21  \\
        LLaVA-1.5-13B (40) & 8--13  \\
        InternVL3.5-8B (36) & 4--26 \\
        \hline
    \end{tabular}
    \caption{Grounding-sensitive layers for different MLLMs.}
    \label{tab:grounding_layers}
\end{table}

\end{document}